\def\year{2021}\relax
\pdfoutput=1
\documentclass[letterpaper]{article} 
\usepackage{aaai21}  
\usepackage{times}  
\usepackage{helvet} 
\usepackage{courier}  
\usepackage[hyphens]{url}  
\usepackage{graphicx} 
\usepackage{graphicx}  
\usepackage{natbib}  
\usepackage{caption} 
\usepackage{bm}
\usepackage{amsmath, amssymb, amsfonts}
\usepackage{enumitem}
\title{Making the Relation Matters: Relation of Relation Learning Network for \\Sentence Semantic Matching}
\author{
		Kun Zhang\textsuperscript{\rm \dag},
		Le Wu\textsuperscript{\rm \dag$\ddagger$},
		Guangyi Lv\textsuperscript{\rm \S},
		Meng Wang\textsuperscript{\rm \dag$\ddagger$}\thanks{Corresponding author},
		Enhong Chen\textsuperscript{\rm \S},
		Shulan Ruan\textsuperscript{\rm \S}\\
}
\affiliations{
	\textsuperscript{\rm \dag}School of Computer Science and Information Engineering, Hefei University of Technology\\
	\textsuperscript{\rm $\ddagger$} Institute of Artificial Intelligence, Hefei Comprehensive National Science Center \\
	\textsuperscript{\rm \S}Anhui Province Key Laboratory of Big Data Analysis and Application, University of Science and Technology of China \\
	\{zhang1028kun, lewu.ustc, eric.mengwang\}@gmail.com, 
	\{gylv,slruan\}@mail.ustc.edu.cn, cheneh@ustc.edu.cn
}

\begin{document}

\maketitle

\newcommand{\shortname}{\emph{R$^2$-Net}}
\newcommand{\fullname}{\emph{Relation of Relation Learning Network (R$^2$-Net)}}

\begin{abstract}
	Sentence semantic matching is one of the fundamental tasks in natural language processing, which requires an agent to determine the semantic relation among input sentences. 
	Recently, deep neural networks have achieved impressive performance in this area, especially BERT. 
	Despite their effectiveness, most of these models treat output labels as meaningless one-hot vectors, underestimating the semantic information and guidance of relations that these labels reveal, especially for tasks with a small number of labels. 
	To address this problem, we propose a \fullname~for sentence semantic matching. 	
	Specifically, we first employ BERT to encode the input sentences from a global perspective.
	Then a CNN-based encoder is designed to capture keywords and phrase information from a local perspective. 
	To fully leverage labels for better relation information extraction, we introduce a self-supervised relation of relation classification task for guiding \shortname~to consider more about relations. 
	Meanwhile, a triplet loss is employed to distinguish the intra-class and inter-class relations in a finer granularity.
	Empirical experiments on two sentence semantic matching tasks demonstrate the superiority of our proposed model. 
	As a byproduct, we have released the codes to facilitate other researches.
\end{abstract}

\section{Introduction}
\begin{figure}
	\centering
	\includegraphics[width=0.35\textwidth]{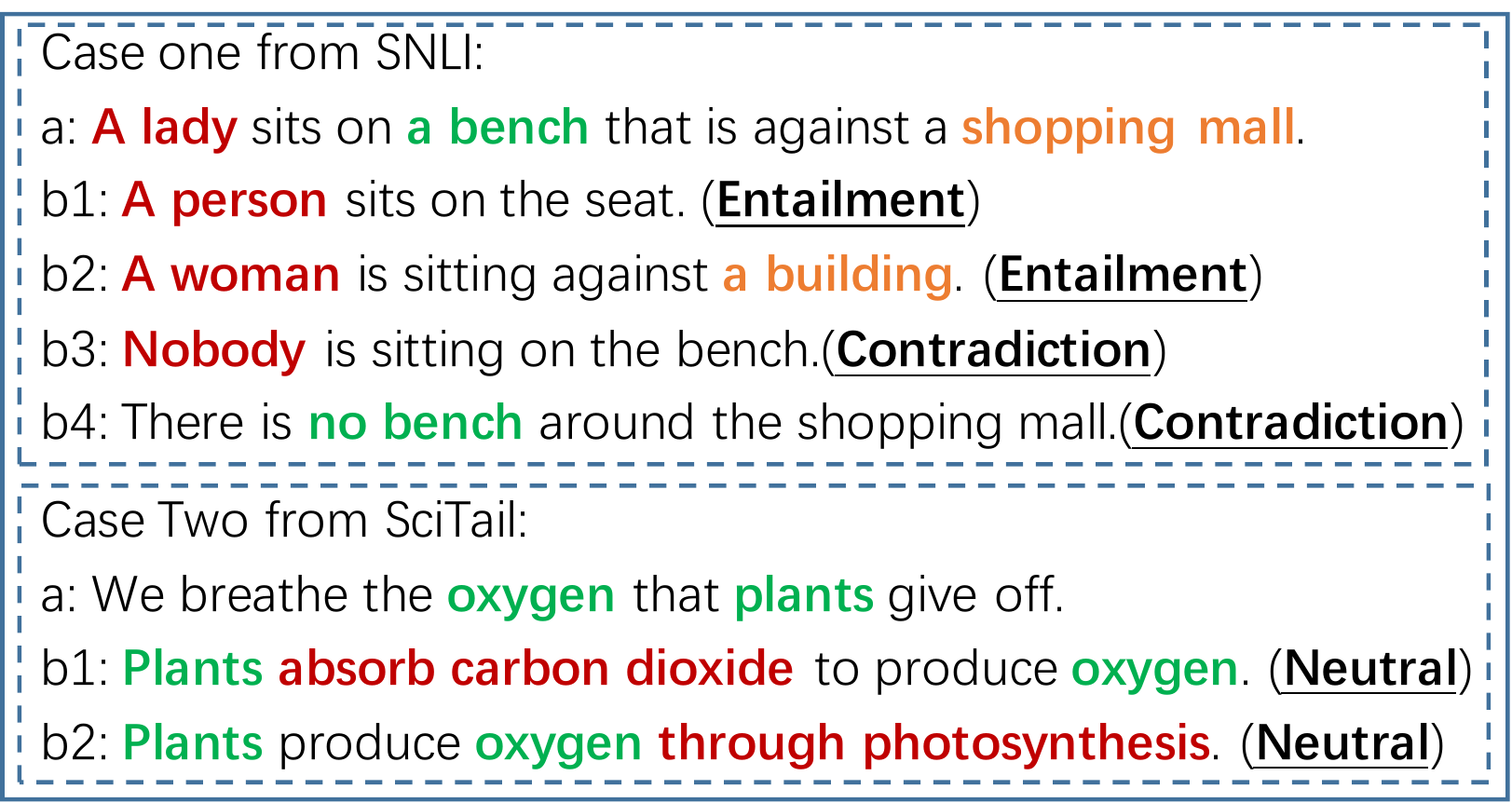}
	\caption{Some examples from SNLI and SciTail datasets.}
	\label{f:example}
\end{figure}

Sentence semantic matching is a fundamental \textit{Natural Language Processing~(NLP)} task that tries to infer the most suitable label for a given sentence pair. For example, Natural Language Inference~(NLI) targets at classifying the input sentence pair into one of the three relations~(i.e., \textit{Entailment, Contradiction, Neutral})~\cite{Kim2018SemanticSM}. Paraphrase Identification~(PI) aims at identifying whether the input sentence pair expresses the same meaning~\cite{dolan2005automatically}. Figure~\ref{f:example} gives some examples with different semantic relations from different datasets.

As a fundamental technology, sentence semantic matching has been applied successfully into many NLP fields, e.g., information retrieval~\cite{Clark2016CombiningRS}, question answering~\cite{liu2018finding}, and dialog system~\cite{serban2016building}. 
Currently, most work leverages the advancement of representation learning techniques~\cite{devlin2018bert,vaswani2017attention} to tackle this task. 
They focus on input sentences and design different architectures to explore sentence semantics comprehensively and precisely. 
Among all these methods, BERT~\cite{devlin2018bert} plays an important role. 
It adopts multi-layer transformers to make full use of large corpus~(i.e., BooksCorpus and English Wikipedia) for the powerful pre-trained model. 
Meanwhile, two self-supervised learning tasks~(i.e., Masked LM and Next Sentence Prediction) are designed to better analyze sentence semantics and capture as much information as possible. 
Based on BERT, plenty of work has made a big step in sentence semantic modeling~\cite{liu2019multi,Radford2018ImprovingLU}.

In fact, since relations are the predicting targets of sentence semantic matching task, most methods do not pay enough attention to the relation learning. 
They just leverage annotated labels to represent relations, which are formulated as one-hot vectors. 
However, these independent and meaningless one-hot vectors cannot reveal the rich semantic information and guidance of relations~\cite{zhang-etal-2018-multi}, which will cause an information loss. 
\citeauthor{gururangan2018annotation}~(\citeyear{gururangan2018annotation}) has observed that different relations among sentence pairs imply specific semantic expressions. 
Taking Figure~\ref{f:example} as an example, most sentence pairs with ``\textit{contradiction}'' relation contain negation words~(e.g., \textit{nobody, never}). 
``\textit{entailment}'' relation often leads to exact numbers being replaced with approximates~(\textit{person, some}). 
``\textit{Neutral}'' relation will import some correct but irrelevant information~(e.g., \textit{absorb carbon dioxide}). 
Moreover, the expressions between sentence pairs with different relations are very different. 
Therefore, the comparison and contrastive learning among different relations~(e.g., pairwise relation learning) can help models to learn more about the semantic information implied in the relations, which in turn helps to strengthen the sentence analysis ability of models.
They should be treated as more than just meaningless one-hot vectors.

One of the solutions for better relation utilization is the embedding method inspired by Word2Vec. 
Some researchers try to jointly encode the input sentences and labels in the same embedding space for better relation utilization during sentence semantic modeling~\cite{du2019explicit,wang-etal-2018-joint-embedding}. 
Despite the progress they have achieved, label embedding method requires more data and parameters to achieve better utilization of relation information. 
It still cannot fully explore the potential of relations due to the small number of relation categories or the lack of explicit label embedding initialization~\cite{wang-etal-2018-joint-embedding}.

To this end, in this paper, we propose a novel \fullname~approach to make full use of relation information in a simple but effective way. 
In concrete details, we first utilize pre-trained BERT to model semantic meanings of the input words and sentences from a global perspective. 
Then, we develop a CNN-based encoder to obtain partial information~(\textit{keywords and phrase information}) of sentences from a local perspective. 
Next, inspired by self-supervised learning methods in BERT training processing, we propose a \textbf{R}elation of \textbf{R}elation~(R$^2$) classification task to enhance the learning ability of \shortname~for the implicit common features corresponding to different relations.   
Moreover, a triplet loss is used to constrain the model, so that the intra-class and inter-class relations are analyzed better.  
Along this line, input sentence pairs with the same relations will be represented much closer and vice versa further apart. 
Relation information is properly integrated into sentence pair modeling processing, which is in favor of tackling the above challenges and improving the model performance. 
Extensive evaluations of two sentence semantic matching tasks (i.e., NLI and PI) demonstrate the effectiveness of our proposed \shortname~and its advantages over state-of-the-art sentence semantic matching baselines.

\section{Related Work}
In this section, we mainly introduce the related work from two aspects: 1) \textit{Sentence Semantic Matching}, and 2) \textit{Label Embedding for Text Classification}.

\subsection{Sentence Semantic Matching}
With the development of various neural network technologies such as CNN~\cite{kim2014convolutional}, GRU~\cite{Chung2014EmpiricalEO}, and the growing importance of the attention mechanism~\cite{vaswani2017attention,Parikh2016ADA}, plenty of methods have been exploited for sentence semantic matching on large datasets like SNLI~\cite{bowman2015large}, SciTail~(Khot et al.~\citeyear{khot2018scitail}), and Quora~(Iyer et al.~\citeyear{iyer2017first}). 
Traditionally, researchers try to fully use neural network technologies to model semantic meanings of sentences in an end-to-end fashion. 
Among them, CNNs focus on the local context extraction with different kernels, and RNNs are mainly utilized to capture the sequential information and semantic dependency. 
For example, \citeauthor{mou2015natural}~(\citeyear{mou2015natural}) employed a tree-based CNN to capture the local context information in sentences. 
\citeauthor{zhang2018ImageEnhance}~(\citeyear{zhang2018ImageEnhance}) combined CNN and GRU into a hybrid architecture, which utilizes the advantages of both networks. They used CNN to generate phrase-level semantic meanings and GRU to model the word sequence and dependency between sentences.

Recently, attention-based methods have shown very promising results on many NLP tasks, such as machine translation~(Bahdanau et al.~\citeyear{Bahdanau2014NeuralMT}), reading comprehension~\cite{zheng2019human}, and NLI~\cite{bowman2016fast}. 
Attention helps to extract the most important parts in sentences, capture semantic relations, and align the elements of two sentences properly~\cite{Cho2015DescribingMC,zhang2017context}.
It has become an essential component for improving model performance and sentence understanding. 
Early attempts focus on designing different attention methods that are suitable for specific tasks, like inner-attention~\cite{Liu2016LearningNL}, co-attention~\cite{Kim2018SemanticSM}, and multi-head attention~\cite{shen2017disan}.
To fully explore the potential of attention mechanism, \citeauthor{zhang2019drr}~(\citeyear{zhang2019drr}) proposed a dynamic attention mechanism, which imitates human reading behaviors to select the most important word at each reading step. This method has achieved impressive performance. 
Another direction is pre-trained methods. \citeauthor{devlin2018bert}~(\citeyear{devlin2018bert}) used very large corpus and multi-layer transformers to obtain a powerful per-trained BERT. 
This method leverages multi-head self-attention to encode sentences and achieves remarkable performances on various NLP tasks. With the powerful representation ability, pre-trained BERT 
model has accelerated the NLP research. 

However, most of these methods only focus on the input sentences and treat the labels as meaningless one-hot vectors, which ignores the potential of label information~\cite{zhang-etal-2018-multi}. There still remains plenty of space for further improvement on sentence semantic matching.

\subsection{Label Embedding for Text Classification} 
As an extremely important part of training data, labels contain much implicit information that needs to be explored. 
In computer vision, researchers have proposed label embedding methods to make full use of label information. 

However, research on explicit label utilization in NLP is still a relatively new domain. 
One possible reason is that there are not that many labels in NLP tasks. Thus, label information utilization is only considered on the task with relatively a large number of labels or multi-task learning. 
For example, \citeauthor{zhang-etal-2018-multi}~(\citeyear{zhang-etal-2018-multi}) proposed a multi-task label embedding method for better implicit correlations and common feature extraction among related tasks. 
\citeauthor{du2019explicit}~(\citeyear{du2019explicit}) designed an explicit interaction model to analyze the fine-grained interaction between word representations and label embedding. 
They have achieved impressive performance on text classification tasks.  
In addition,  \citeauthor{wang-etal-2018-joint-embedding}~(\citeyear{wang-etal-2018-joint-embedding}) and \citeauthor{pappers2019gile}~(\citeyear{pappers2019gile}) transferred the text classification task to a label-word joint embedding problem. 
They leveraged the semantic vectors of labels to guide models to select the important and relevant parts of input sentences for better performance. 
The above work demonstrates the superiority of explicit label utilization and inspires us to make better use of label information.

\begin{figure*}
	\centering
	\includegraphics[width=0.86\textwidth]{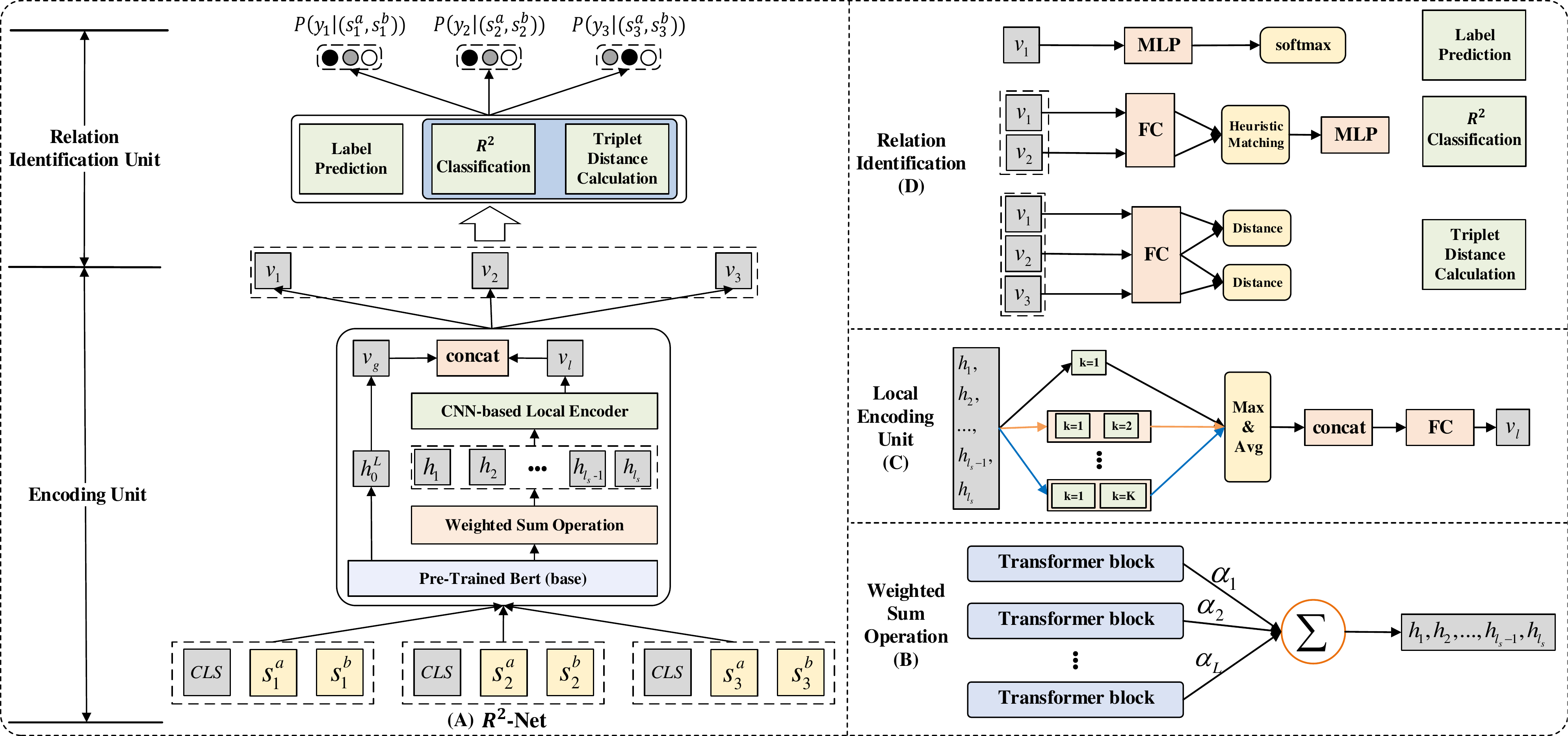}
	\caption{Architecture of \fullname.} 
	\label{f:model}
\end{figure*}

\section{Problem Statements}
\label{s:problem}

In this section, we will introduce the definition of sentence semantic matching task and our proposed relation of relation classification task. 

\subsection{Sentence Semantic Matching}
Sentence semantic matching task can be formulated as a supervised classification. 
Given two input sentences $\bm{s}^a = \{\bm{x}_1^a, \bm{x}_2^a, ..., \bm{x}_{l_a}^a \}$ and $\bm{s}^b = \{\bm{x}_1^b, \bm{x}_2^b, ..., \bm{x}_{l_b}^b \}$, where $\bm{x}_i^a$ and $\bm{x}_j^b$ are feature tokens for each sentence. 
The goal of this task is to train a classifier $\xi$, which is capable of computing the conditional probability $P(y|\bm{s}^a, \bm{s}^b)$ and predicting the relation for input sentence pair based on the probability. 
\begin{equation}
	\label{eq:task}
	\begin{split}
		&P(y|\bm{s}^a, \bm{s}^b)= \xi(\bm{s}^a, \bm{s}^b), \\ 
		&y^{*} =argmax_{y\in\mathcal{Y}}P(y|\bm{s}^a, \bm{s}^b),
	\end{split}
\end{equation} 
where the true label $y\in\mathcal{Y}$ indicates the semantic relation between the input sentence pair. 
$\mathcal{Y}=\{entailment, contradiction, neutral\}$ for NLI task and $\mathcal{Y}=\{Yes, No\}$ for PI task.

\subsection{Relation of Relation Classification}

\citeauthor{gururangan2018annotation}~(\citeyear{gururangan2018annotation}) has observed that relations can be helpful to reveal some implicit features or patterns for semantic understanding and matching. 
In order to properly and fully utilize relation information, we propose a \textbf{R}elation of \textbf{R}elation~(R$^2$) classification task to guide models to understand sentence relation more precisely. 
Given two input sentence pairs $(\bm{s}^a_1, \bm{s}^b_1)$ and $(\bm{s}^a_2, \bm{s}^b_2)$, the goal is to learn a classifying function $\mathcal{F}$ with the ability to identify whether these two input pairs have the same semantic relation:
\begin{equation}
	\label{eq:r2-task}
	\begin{split}
		\mathcal{F}((\bm{s}^a_1, \bm{s}^b_1), (\bm{s}^a_2, \bm{s}^b_2)) = 
		\begin{cases} 
			1,  & \mbox{if } y_1 = y_2, \\
			0, & \mbox{if } y_1 \not= y_2,
		\end{cases}
	\end{split}
\end{equation} 
where $y_1$ and $y_2$ stand for the semantic relations of two input sentence pairs, respectively. 

In order to make full use of relation information and do better sentence semantic matching, the following important questions should be considered:
\begin{itemize}\setlength{\itemsep}{0pt}
	\item Since relations are the predicting targets, how to make full use of relation information to improve model performance properly without leaking it?
	\item How to integrate R$^2$ task into matching task effectively for relation usage and performance improvement?
\end{itemize}

To this end, we propose \shortname~to properly and fully utilize relation information, and tackle the above issues. Next, we will introduce the technical details of \shortname.

\section{Relation of Relation Learning Network (R$^2$-Net)}
\label{s:model}

The overall architecture of \shortname~is shown in Figure~\ref{f:model}(A).
To better describe how \shortname~tackles the above tasks and integrates R$^2$ task to enhance the model ability on sentence semantic matching, similar to Section~\ref{s:problem}, we also elaborate the technical details from two aspects: 1) Sentence Semantic Matching Part; 2) Relation of Relation Learning Part. 

\subsection{Sentence Semantic Matching Part}
\label{s:encode_unit}
This part focuses on identifying the most suitable label for a given input sentence pair. 
Specifically, for an input sentence pair, we first utilize powerful BERT to generate sentence semantic representation globally. 
Meanwhile, we develop a CNN-based encoder to capture the keywords and phrase information from a local perspective. 
Thus, the input sentence pair can be encoded in a comprehensive manner. 
Based on the comprehensive representation, we leverage a multi-layer perceptron to predict the corresponding label. 

\subsubsection{Global Encoding.}
With the full usage of large corpus and multi-layer transformers, BERT~\cite{devlin2018bert} has accomplished much progress in many NLP tasks. 
Thus, we select BERT to generate sentence semantic representations for the input. 
Moreover, inspired by ELMo~\cite{Peters2018DeepCW}, we also use the weighted sum of all the hidden states of words from different transformer layers as the final contextual representations of input words in sentences. 

Specifically, we first split the input sentence pair $(\bm{s}^a, \bm{s}^b)$ into BPE  tokens~(Sennrich et al.~\citeyear{sennrich2015neural}). 
Then, we concatenate two sentences to the required format, in which ``[\textit{SEP}]'' is adopted to concatenate two sentences and ``[\textit{CLS}]'' is added at the beginning and the end of the whole sequence. 
Then, we use multi-layer transformer blocks to obtain the representations of words and sentences in the input. 
Moreover, as illustrated in Figure~\ref{f:model}(B), suppose there are $L$ layers in the BERT. The contextual word representations in the input sentence pairs is then a pre-layer weighted sum of transformer block output, with the weights $\alpha_1, \alpha_2,...,\alpha_L$.
\begin{equation}
	\label{eq:global-encoding}
	\begin{split}
		\bm{h}_0^l, \bm{H}^l &= TransformerBlock(\bm{s}^a, \bm{s}^b), \\
		\bm{H} &= \sum_{l=1}^L\alpha_l\bm{H}^l, \quad \bm{v}_g = \bm{h}_0^L, \\
	\end{split}
\end{equation} 
where $\bm{h}_0^l$ denotes the representation of first token ``[\textit{CLS}]'' at the $l^{th}$ layer, and $\bm{v}_g$ denotes the global semantic representation of the input. 
$\bm{H}^l$ represents the sequence features of the whole input. $\alpha_l$ is the weight of the $l^{th}$ layer in BERT and will be learned during model training.

\subsubsection{Local Encoding.}
The semantic relation within the sentence pair is not only connected with the important words, but also affected by the local information~(e.g., phrase and local structure). 
Though Bert leverages multi-layer transformers to perceive important words to the sentence pair, it still has some weaknesses in modeling local information. 
To alleviate these shortcomings, we develop a CNN-based local encoder to extract the local information from the input.  

Figure~\ref{f:model}(C) illustrates the structure of this local encoder. 
The input of this encoder is the output features $\bm{H}$ from global encoding. 
We use convolution operations with different composite kernels~(e.g., bigram and trigram) to process these features. 
Each operation with different kernels is capable of modeling patterns with different sizes~(e.g., \textit{new couple, tall person}). 
Thus, we can obtain robust and abstract local features of the input sentence pair. 
Next, we leverage average pooling and max pooling to enhance these local features and concatenate them before sending them to a non-linear transformation. 
Suppose we have $K$ different kernel sizes, this process can be formulated as follows:
\begin{equation}
	\label{eq:local-encoding}
	\begin{split}
		\bm{H}^k &= CNN_k(\bm{H}), \quad k=1,2,...,K,\\
		\bm{h}^k_{max} &= max(\bm{H}^k), \bm{h}^k_{avg} = avg(\bm{H}^k), \\
		\bm{h}_{concat} &= [\bm{h}^1_{max};\bm{h}^1_{avg};...;\bm{h}^K_{max};\bm{h}^K_{avg}], \\
		\bm{v}_l &= ReLu(\bm{W}\bm{h}_{concat} + \bm{b}), \\
	\end{split}
\end{equation}
where $CNN_k$ denotes the convolution operation with the $k^{th}$ kernel. $[\cdot;\cdot]$ is the concatenation operation. $\bm{v}_l$ represents the local semantic representation of the input. $\{\bm{W}, \bm{b}\}$ are trainable parameters. $ReLu(\cdot)$ is the activation function. 

After getting the global representation $\bm{v}_g$ and local representation $\bm{v}_l$, we investigate the different fusion methods to integrate them together, including simple concatenation, weighed concatenation, as well as weighted sum. Finally, we obtain that simple concatenation is flexible and can achieve comparable performance without adding more training parameters. 
Thus, we employ the concatenation $\bm{v} = [\bm{v}_g;\bm{v}_l]$ as the final semantic representation of the input sentence pair.

\begin{table*}
	\centering
	\caption{Performance (accuracy) of models on different NLI dataset.}
	\begin{footnotesize}
		\begin{tabular}{lccc} \hline
			\textbf{Model} & \textbf{Full test} & \textbf{Hard test} & \textbf{SICK test} \\ \hline
			(1) CENN~\cite{zhang2017context} & 82.1\% & 60.4\% & 81.8\% \\
			(2) CAFE~\cite{Tay2017ACA} & 85.9\% & 66.1\% & 86.1\% \\
			(3) Gumbel TreeLSTM~\cite{choi2018learning} & 86.0\% & 66.7\% & 85.8\%\\
			(4) Distance-based SAN~\cite{im2017distance} & 86.3\% & 67.4\% & 86.7\% \\
			(5) DRCN~\cite{Kim2018SemanticSM} & 86.5\%& 68.3\% & 87.4\% \\ \hline
			(6) DRr-Net~\cite{zhang2019drr} & 87.5\% & 71.2\% & 87.8\% \\ 
			(7) Dynamic Self-Attention~\cite{yoon2018dynamic} & 87.4\% & 71.5\% & 87.7\% \\ 
			(8) Bert-base~\cite{devlin2018bert} & 90.3\% & 80.8\% & 88.5\% \\ \hline
			(9) \shortname & \textbf{91.1\%} & \textbf{81.0\%} & \textbf{89.2\%} \\ 
			\hline
		\end{tabular}
	\end{footnotesize}
	\label{t:snli-result}
\end{table*}

\subsubsection{Label Prediction.}
This component is adopted to predict the label of input sentence pair, which is an essential part of traditional sentence semantic matching methods. 
To be specific, the input of this component is the semantic representation $\bm{v}$. We leverage a two-layer MLP to make the final classification, which can be formulated as follows:
\begin{equation}
	\label{eq:label-prediction}
	\begin{split}
		P(y|(\bm{s}^a, \bm{s}^b)) = MLP_1(\bm{v}). \\
	\end{split}
\end{equation} 

\subsection{Relation of Relation Learning Part}
\label{s:predict_unit}
This part aims at properly and fully using relation information of input sentence pairs to enhance the model performance on sentence semantic matching. 
In order to achieve this goal, we employ two critical modules to analyze the \textit{pairwise relation} and \textit{triplet based relation} simultaneously. 
Next, we will describe each module in detail.

\subsubsection{Relation of Relation Classification. }
\label{s:r2_unit}
Inspired by self-supervised learning methods in BERT, we intend \shortname~to make full use of relation information among input sentence pairs in a similar way. 
Therefore, we introduce R$^2$ classification task into sentence semantic matching. Instead of just identifying the most suitable relation of input sentence pairs, 
we plan to obtain more knowledge about the input sentence pair by analyzing the \textit{pairwise relation} between the semantic representations ($\bm{v}_1$ for pair $(\bm{s}^a_1, \bm{s}^b_1)$, and $\bm{v}_2$ for pair $(\bm{s}^a_2, \bm{s}^b_2)$). 
Since a learnable nonlinear transformation between representations and loss substantially improves the model performance~\cite{Chen2020ASF}, we first transfer $\bm{v}_1$ and $\bm{v}_2$ with a nonlinear transformation. 
Then, we leverage heuristic matching~\cite{Chen-Qian2017ACL} to model the similarity and difference between $\bm{v}_1$ and $\bm{v}_2$. 
Next, we send the result $\bm{u}$ to a MLP with one hidden layer for final classification. This process is formulated as follows:
\begin{equation}
	\label{eq:r2-prediction}
	\begin{split}
		&\bar{\bm{v}}_1 = ReLu(\bm{W}_r\bm{v}_1 + \bm{b}_r), \\
		&\bar{\bm{v}}_2 = ReLu(\bm{W}_r\bm{v}_2 + \bm{b}_r), \\
		&\bm{u} = [\bar{\bm{v}}_1; \bar{\bm{v}}_2; (\bar{\bm{v}}_1 \odot \bar{\bm{v}}_2); (\bar{\bm{v}}_1 - \bar{\bm{v}}_2)], \\
		&P(\hat{y}|(\bm{s}^a_1, \bm{s}^b_1), (\bm{s}^a_2, \bm{s}^b_2)) = MLP_2(\bm{u}), \\
	\end{split}
\end{equation}
where concatenation can retain all the information~\cite{zhang2017context}. The element-wise product is a certain measure of ``similarity'' of two sentences~\cite{mou2016natural}. Their difference can capture the degree of distributional inclusion in each dimension~\cite{weeds2014learning}. 
$\hat{y}\in\{1, 0\}$ indicates whether two input sentence pairs have same relation.

\subsubsection{Triplet Distance Calculation. }
Apart from leveraging R$^2$ classification task to learn pairwise relation information, we also intend to learn intra-class and inter-class information from the \textit{triplet based relation}. 
Thus, we also introduce a triplet loss~(Schroff et al.~\citeyear{Schroff2015FaceNetAU}) into \shortname. 
As a fundamental similarity function, triplet loss is widely applied in information retrieval area~\cite{Liu2010LearningTR}, and is able to reduce the distance of input pairs with the same relation and increase the distance of these with different relations. 
Therefore, we first calculate the corresponding distances in this module. 
To be specific, the inputs of this component are three semantic representations: 
$\bm{v}_a$ for anchor pair $(\bm{s}^a_a, \bm{s}^b_a)$, $\bm{v}_p$ for positive pair $(\bm{s}^a_p, \bm{s}^b_p)$, $\bm{v}_n$ for negative pair $(\bm{s}^a_n, \bm{s}^b_n)$. 
In order to obtain better results, we first transform them into a common space with a full connection layer~\cite{Chen2020ASF}. 
Then, we calculate the distance between anchor and positive pairs, and the distance between anchor and negative pairs, respectively. 
This process is formulated as follows. 
\begin{equation}
	\label{eq:distance-prediction}
	\begin{split}
		&\bar{\bm{v}}_a = ReLu(\bm{W}_d\bm{v}_a + \bm{b}_d), \\
		&\bar{\bm{v}}_p = ReLu(\bm{W}_d\bm{v}_p + \bm{b}_d), \\
		&\bar{\bm{v}}_n = ReLu(\bm{W}_d\bm{v}_n + \bm{b}_d), \\
		& d_{ap} = Dist(\bar{\bm{v}}_a, \bar{\bm{v}}_p), \quad
		d_{an} = Dist(\bar{\bm{v}}_a, \bar{\bm{v}}_n), \\
	\end{split}
\end{equation}
where $\{\bm{W}_d, \bm{b}_d\}$ are trainable parameters. $Dist(\cdot)$ is the distance calculation function.

\section{Experiments}
\label{s:experiment}
In this section, the details about model implementation are firstly presented.
Then, five benchmark datasets on which the model is evaluated are introduced.
Next, a detailed analysis about the model and experimental results is made.

\begin{table}
	\centering
	\caption{Experimental Results~(accuracy) on SciTail dataset.}
	\begin{footnotesize}
		\begin{tabular}{lc} \hline
			\textbf{Model} & \textbf{SciTail test}\\ \hline
			(1) CAFE~\shortcite{Tay2017ACA} & 83.3\% \\
			(2) ConSeqNet~\shortcite{Wang2018ImprovingNL} & 85.2\% \\
			(3) BiLSTM Max-Out~\shortcite{Mihaylov2018CanAS} & 85.4\% \\
			(4) HBMP~(Talman et al.\citeyear{talman2018natural}) & 86.0\% \\ 
			(5) DRr-Net~\shortcite{zhang2019drr} & 87.4\% \\ \hline
			(6) Transformer LM~\shortcite{Radford2018ImprovingLU} & 88.3\% \\ 
			(7) Bert-base~\shortcite{devlin2018bert} & 92.0\% \\ \hline
			(8) \shortname & \textbf{92.9\%} \\
			\hline
		\end{tabular}
	\end{footnotesize}
	\label{t:scitail-result}
\end{table}

\subsection{Training Details}

\subsubsection{Loss Function.}
As is mentioned in Section~\ref{s:problem}, both sentence semantic matching and R$^2$ task can be treated as classification tasks. 
Thus, we employ \textit{Cross-Entropy} as the loss for each input as follows:
\begin{equation}
	\label{eq:loss_classify}
	\begin{split}
		L_{s} &= -\bm{y}_i \mathrm{log} P(y_i | (\bm{s}^a_i, \bm{s}^b_i)), \\ 
		L_{R^2} & = -\hat{\bm{y}}_{i} \mathrm{log} P(\hat{y}|((\bm{s}^a_1, \bm{s}^b_1), (\bm{s}^a_2, \bm{s}^b_2))_i), \\
	\end{split}
\end{equation}
where $\bm{y}_i$ is the one-hot vector for the true label of the $i^{th}$ instance. $\hat{\bm{y}}_{i}$ is the one-hot vector for the true relation of relations of the $i^{th}$ instance pair. 

Moreover, in order to learn more from relations and achieve better performance, we also introduce the triplet loss to force \shortname~to better analyze the intra-class and inter-class information among sentence pairs with same or different relations:
\begin{equation}
	\label{eq:loss_dist}
	\begin{split}
		L_{d} = max((d_{ap}-d_{an}+\alpha)_i, 0), \\
	\end{split}
\end{equation}
where $\alpha$ is the margin. $(\cdot)_i$ denotes the $i^{th}$ triplet pair. 

Since these three loss functions require different number of inputs, we modify the input of \shortname~to have three input sentence pairs~(i.e., anchor pair, positive pair, and negative pair), as shown in Figure~\ref{f:model}(A). 
Then, we calculate $L_{s}^1,L_{s}^2,L_{s}^3$ for label prediction loss of each input pair, randomly sample two groups from the input to calculate $L_{R^2}^1, L_{R^2}^2$ for R$^2$ task loss, and use three input pairs to calculate $L_{d}$ for triplet loss. 
Finally, we treat the weighed sum of these losses with a hyper-parameter $\beta$ as the loss function for entire model as follows:
\begin{equation}
	\label{eq:loss}
	\begin{split}
		L = \frac{1}{N}\sum_{i=1}^{N}(\beta\frac{L_{s}^1+L_{s}^2+L_{s}^3}{3} + (1-\beta)(\frac{L_{R^2}^1+L_{R^2}^2}{2}+ L_{d})). \\
	\end{split}
\end{equation}

\begin{table*}
	\centering
	\caption{Experimental Results~(accuracy) on Quora and MSRP datasets.}
	\begin{footnotesize}
		\begin{tabular}{lcc} \hline
			\textbf{Model} & \textbf{Quora test} & \textbf{MSRP test} \\ \hline
			(1) CENN~\cite{zhang2017context} & 80.7\% & 76.4\%\\
			(2) L.D.C~\cite{Wang2016SentenceSL} & 85.6\% & 78.4\%\\
			(3) REL-TK~(Filice et al.~\citeyear{filice2015structural}) & - & 79.1\% \\
			(4) BiMPM~\cite{Wang2017BilateralMM} & 88.2\% & -\\
			(5) pt-DecAttachar.c~\cite{Tomar2017NeuralPI} & 88.4\% & -\\
			(6) DIIN~\cite{gong2017natural} & 89.1\% & -\\ \hline
			(7) DRr-Net~\cite{zhang2019drr} & 89.8\% & 82.9\% \\
			(8) DRCN~\cite{Kim2018SemanticSM} & 90.2\% & 82.5\% \\ 
			(9) BERT-base~\cite{devlin2018bert} & 91.0\% & 84.2\%\\ \hline
			(10) \shortname & \textbf{91.6\%} & \textbf{84.3\%} \\
			\hline
		\end{tabular}
	\end{footnotesize}
	\label{t:pi-result}
\end{table*}

\subsubsection{Model Implementation. }
We have tuned the hyper-parameters on validation set for best performance, and have used early-stop to select the best model. 
Since \shortname~has different hyper-parameter settings on different datasets, we list some common hyper-parameters as follows. 

We apply the BERT-base with 12 layers, hidden size 768, and 12 heads. 
The kernel sizes of CNN in local encoding unit are $d_k=1, 2, 3$. 
The hidden state size of MLP in \shortname~is $d_m=300$.
The distance we use in distance calculation component is \textit{Euclidean Distance}.
The margin $\alpha$ in the triplet loss is $\alpha=0.2$.
For the pre-trained BERT, we set the learning rate $10^{-5}$ and use AdamW to fine-tune the parameters. 
For the rest of parameters, we set the initial learning rate to be $10^{-3}$ and decrease its value as the model training. 
An Adam optimizer with $\beta_1 =0.9$ and $\beta_2 = 0.999$ is adopted to optimize these parameters. 
The entire model is implemented with PyTorch and Transformers\footnote{https://github.com/huggingface/transformers}, and is trained on two Nvidia Tesla V100-SXM2-32GB GPUs.

\subsection{Data Description}
\label{s:data}
In this section, we give a brief introduction of the datasets on which we evaluate all models. They are as follow:
\begin{itemize}
	\item \textbf{SNLI:} SNLI dataset~\cite{bowman2015large} contains $570,152$ human annotated sentence pairs. 
	The premise sentences are drawn from the captions of Flickr30k corpus~\cite{young2014image}, and the hypothesis sentences are manually composed. 
	Despite the original test set, we also select the challenging hard subset~\cite{gururangan2018annotation} to evaluate the models.
	
	\item \textbf{SICK:} SICK dataset~\cite{marelli2014semeval} contains $10,000$ English sentence pairs, generated from  8K ImageFlickr dataset~(Hodosh et al.~\citeyear{hodosh2013framing}) and STS MSR-video description dataset\footnote{https://www.cs.york.ac.uk/semeval-2012/}. 
	Each sentence pair is generated from randomly selected subsets of the above sources and manually labeled with the label set as SNLI did. 
	
	\item \textbf{SciTail}: SciTail dataset~(Khot et al.~\citeyear{khot2018scitail}) is created from multiple-choice science exams and web sentences. It has $27,026$ examples with $10,101$ \textit{Entailment} examples and $16,925$ \textit{Neutral} examples. 
	
	\item \textbf{Quora}: Quora dataset~(Iyer et al.~\citeyear{iyer2017first}) contains over $400,000$ potential question duplicate pairs, which are drawn from Quora website\footnote{https://www.quora.com/}. 
	This dataset has balanced positive and negative labels, indicating whether the line truly contains a duplicate pair.
	
	\item \textbf{MSRP}: MSRP dataset~\cite{dolan2005automatically} consists of $5,801$ sentence pairs with a binary label. 
	The sentences are distilled from a database of $13,127,938$ sentence pairs, extracted from $9,516,684$ sentences in $32,408$ news clusters from the web. 
\end{itemize}

\begin{figure*}
	\centering
	\includegraphics[width=0.84\textwidth]{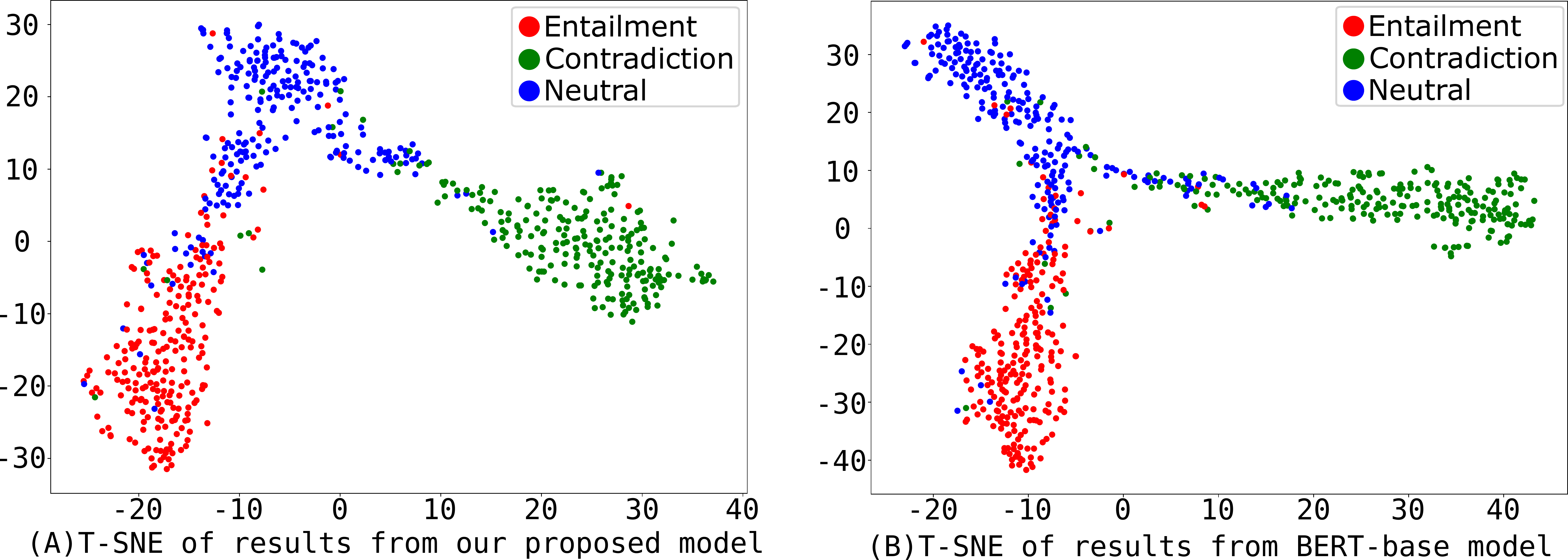}
	\caption{Visualization of representation $\bm{v}$ from \shortname~and BERT-base models.} 
	\label{f:case}
\end{figure*}

\subsection{Experimental Results}
In this section, we will give a detailed analysis about models and experimental results. We have to note that we use \textit{accuracy} on different test sets to evaluate the model performance. 

\subsubsection{Performance on NLI task.}
We compared our proposed \shortname~to several published state-of-the-art baselines on different NLI datasets. All results are summarized in Table~\ref{t:snli-result} and Table~\ref{t:scitail-result}. Several observations are presented as follows.
\begin{itemize}
	\item It is clear that \shortname~achieves highly comparable performance over all the datasets: SNLI, SICK, and SciTail. Specifically, \shortname~first fully uses BERT and CNN-based encoder to get a comprehensive understanding of sentence semantics from global and local perspectives. This is one of the reasons that \shortname~outperforms other BERT-free baselines by a large margin. Another important reason is that \shortname~employs R$^2$ task and triplet loss to make full use of relation information. Along this line, \shortname~is capable of obtaining intra-class and inter-class knowledge among sentence pairs with the same or different relations. Thus, it can achieve better performance than all baselines, including the BERT-base model.
	
	\item \shortname~has more stable performance on the challenging NLI hard test, in which the pairs with obvious identical words are removed~\cite{gururangan2018annotation}. Despite the obvious indicators, these still have implicit patterns for relations among sentence pairs. By considering R$^2$ task and triplet loss, \shortname~has the ability to fully use relation information and obtain the implicit information, which leads to a better performance.
	
	\item BERT-base model~\cite{devlin2018bert} outperforms other BERT-free baselines by a large margin. The main reasons can be grouped into two parts. First, BERT takes advantages of multi-layer transformers to learn sentence patterns and sentence semantics on a large corpus. Second, BERT adopts two self-supervised learning tasks~(i.e., MLM and NSP) to better analyze the important words within a sentence and semantic connection between sentences. However, BERT still focuses on the input sequence, underestimating the rich semantic information that relations imply. Therefore, its performance is not as good as that \shortname~achieves.
	
	\item Among BERT-free baselines, DRr-Net~\cite{zhang2019drr} and dynamic self-attention~\cite{yoon2018dynamic} achieve impressive performances.  
	First, their performance proves that multi-layer structure and CNN have better ability to model sentence semantics from global and local perspectives. 
	Then, they all develop a dynamic attention mechanism to improve self-attention mechanism. 
	However, their encoding capability of extracting features or generating semantic representations is still not comparable with BERT. 
	This observation inspires us that using powerful BERT as a basic encoder will be a better choice.
\end{itemize}

\begin{table}
	\centering
	\caption{Ablation performance (accuracy) of \shortname.}
	\begin{footnotesize}
		\begin{tabular}{lcc} \hline
			\textbf{Model} & \textbf{SNLI test} & \textbf{SciTail test} \\ \hline
			(1)~Bert-base & 90.3\% & 92.0\%  \\ \hline
			(2)~\shortname~(w/o local encoder) & 90.7\% & 92.6\%  \\ 
			(3)~\shortname~(w/o R$^2$ task learning) & 90.5\% & 92.3\%  \\ 
			(4)~\shortname~(w/o triplet loss) & 90.9\% & 92.6\%  \\ \hline
			(5)~\shortname~ & \textbf{91.1}\%  & \textbf{92.9}\% \\ 
			\hline
		\end{tabular}
	\end{footnotesize}
	\label{t:ablation-result}
\end{table}

\subsubsection{Performance on PI task.}
Apart from NLI task, we also select PI task to evaluate the model performance. PI task concerns whether two sentences express the same meaning and has broad applications in question answering communities\footnote{https://www.quora.com/}~\footnote{https://www.zhihu.com/}. 
Table~\ref{t:pi-result} reports the performance of models on different datasets. We also list the observations as follows:

\begin{itemize}
	\item \shortname~still achieves highly competitive performance over other baselines. The results demonstrate that R$^2$ task and triplet loss is effective in helping our proposed model to learn more about relations and improve the model performance, even if the number of relations is small. 
	
	\item Almost all models have a better performance on Quora dataset than MSRP dataset. One possible reason is that Quora dataset has more data than MSRP dataset~(over 400k sentences pairs v.s. 5,801 sentence pairs). In addition to the data size, inter-sentence interaction is probably another reason. 
	Lan et al.~(\citeyear{lan2018neural}) observes that Quora dataset contains many sentence pairs with less complicated interactions~(many identical words in sentence pairs). 
	Meanwhile, \shortname~also achieves better improvement on Quora dataset, indicating that more data or better label utilization method is needed for further performance improvement on MSRP dataset. 
\end{itemize}

\subsubsection{Ablation Performance.}
The overall experiments have proved the superiority of \shortname. 
However, which part plays a more important role in performance improvement is still unclear. 
Therefore, we perform an ablation study to verify the effectiveness of each part, including \textit{CNN-based local encoder}, \textit{R$^2$ task classification}, and \textit{triplet loss}. 
The results are illustrated in Table~\ref{t:ablation-result}. 
Note that we select BERT-base as the baseline to compare the importance of each part.
According to the results, we can observe varying degrees of model performance decline. Among all of them, R$^2$ task has the biggest impact, and triple loss has a relatively small impact on the model performance. These observations prove that  R$^2$ task is more important for relation information utilization.

\subsubsection{Case Study. }
To provide some intuitionistic examples for explaining why our model gains a better performance than other baselines, we sample 700 sentence pairs from SNLI dataset and send them to \shortname~and BERT-base models to generate the semantic representation $\bm{v}$.
Then, we leverage t-sne~\cite{maaten2008visualizing} to visualize these representations with the same parameter settings.
Figure~\ref{f:case}(A)-(B) report the results of \shortname~and BERT-base models, respectively. 
By comparing two figures, we can obtain that the representations generated by \shortname~have closer inter-class distances. 
Moreover, the representations have more obvious distinctions between different classes. 
These observations not only explain why our proposed \shortname~achieves impressive performance, but also demonstrates that proper usage of relation information is able to guide models to analyze sentence semantics more comprehensively and precisely, which is in favor of tackling sentence semantic matching.

\section{Conclusion}
In this paper, we presented a simple but effective method named \shortname~for sentence semantic matching. 
This method not only uses powerful BERT and CNN to encode sentences from global and local perspectives, but also makes full use of relation information for better performance enhancement. 
Specifically, we design a R$^2$ classification task to help \shortname~for learning the implicit common knowledge from the pairwise relation learning processing. 
Moreover, a triplet loss is employed to constrain \shortname~for better triplet based relation learning and intra-class and inter-class information analyzing. 
Extensive experiments on NLI and PI tasks demonstrate the superiority of \shortname. In the future, we plan to combine the advantages of label embedding method for better sentence semantic comprehension.

\section{Acknowledgments}


This research was partially supported by grants from the National Natural Science Foundation of China (Grant No. 62006066, 61725203, 61972125, 61732008, and U19A2079), the Major Program of the National Natural Science Foundation of China (91846201), the Fundamental Research Funds for the Central Universities, HFUT. The author wants to thank Cui Gongrongxiu for constructive criticism on the earlier draft.

\small
\bibliographystyle{aaai21}
\bibliography{myReference}

\end{document}